\documentclass[10pt,twocolumn,letterpaper]{article}

\usepackage[pagenumbers]{cvpr} 

\definecolor{lightblue}{rgb}{0.2, 0.5, 0.9}

\definecolor{cvprblue}{rgb}{0.21,0.49,0.74}
\usepackage[pagebackref,breaklinks,colorlinks,allcolors=cvprblue]{hyperref}
\usepackage{multirow}
\usepackage{makecell}
\usepackage{tcolorbox}
\usepackage{amssymb}
\usepackage{svg}
\usepackage{xspace}

\usepackage{booktabs}
\usepackage{graphicx}
\usepackage{array}
\usepackage{amsmath}
\usepackage{pgfplots}

\usepackage{makecell}
\usepackage{bbding}
\usepackage[normalem]{ulem}

\newcommand{\cmark}{\Checkmark}
\newcommand{\xmark}{\XSolidBrush}

\pgfplotsset{compat=1.18}
\usepgfplotslibrary{groupplots,fillbetween}
\usetikzlibrary{arrows.meta,positioning}
\usepackage[capitalize]{cleveref}

\newcommand{\ourmodel}{RECAP-Forcing\xspace}

\def\paperID{***} 
\def\confName{CVPR}
\def\confYear{2027}

\newcommand{\acronymletter}[1]{%
  \raisebox{0.05ex}{\uline{\fontsize{1em}{1em}\selectfont #1}}%
}

\title{
\acronymletter{RECAP}-Forcing:
\acronymletter{RE}taining
\acronymletter{C}ontent
\acronymletter{AP}pearances
for Long Video Generation
}

\author{%
  Haiyang Xu \qquad
  Zheng Ding$^{\dagger}$ \qquad
  Zhuowen Tu \\
  UC San Diego \\
  $^{\dagger}$Project Lead
}

\begin{document}

\twocolumn[{
    \maketitle
    \begin{center}
    \captionsetup{type=figure}
    \vspace{-1.5em}
    \includegraphics[width=\linewidth]{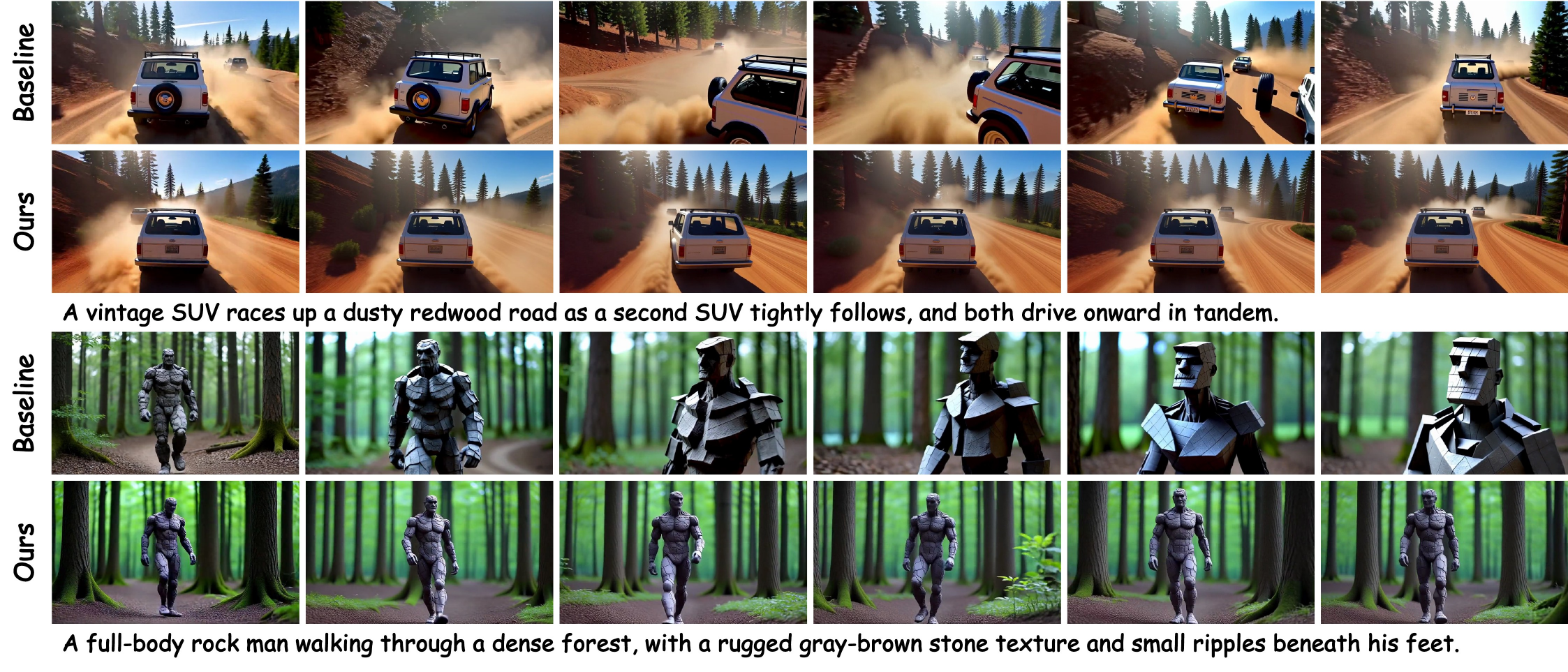}
    \captionof{figure}{\textbf{Minute-long generation, consistent and dynamic.} Two pairs of one-minute videos generated by the same base model, without (Baseline) and with (Ours) \ourmodel applied training-free at inference. As the minute unfolds, the baseline gradually loses content identity and begins to drift and degrade: the trailing SUV dissolves into dust and duplicates, while the rock man fractures and veers off his path. In contrast, \ourmodel preserves subject identity and visual consistency, without sacrificing motion dynamics.}
    \label{fig:teaser}
    \end{center}
}]

\begin{abstract} 
Long autoregressive video generation faces a fundamental memory challenge: with a finite attention window, a model must decide which information from an ever-expanding history to retain. Existing methods organize memory temporally, preserving recent frames while compressing or discarding older ones. We instead propose \ourmodel, organizing memory by appearance novelty. A long video is not merely a sequence of frames, but an evolving cast of subjects, objects, and scenes whose identities must remain consistent over time. We organize memory by retaining the KV cache associated with newly appearing content—such as entering subjects, disoccluded regions, and newly introduced scenes—at the moment it first becomes visible, prioritizing novelty over recency. \textbf{Memory should scale with the amount of newly introduced content, rather than with video length.} This appearance-indexed memory makes long-range consistency an explicit property of the memory structure. Our framework unifies two mechanisms under this single principle. At the beginning of a video, when all visible content is novel, an attention sink preserves the initial scene. As the video evolves, an optical-flow-based novelty bank extends the same principle by selectively retaining newly revealed content. As a training-free inference method with no additional learnable parameters, \ourmodel consistently improves visual quality and semantic fidelity across multiple strong baselines and outperforms existing memory methods.
Project page: \url{https://xxuhaiyang.github.io/RECAP-Forcing/}
\end{abstract}

\begin{figure*}[!t]
\centering
\includegraphics[width=\linewidth, trim={1em 1em 1em 1em}]{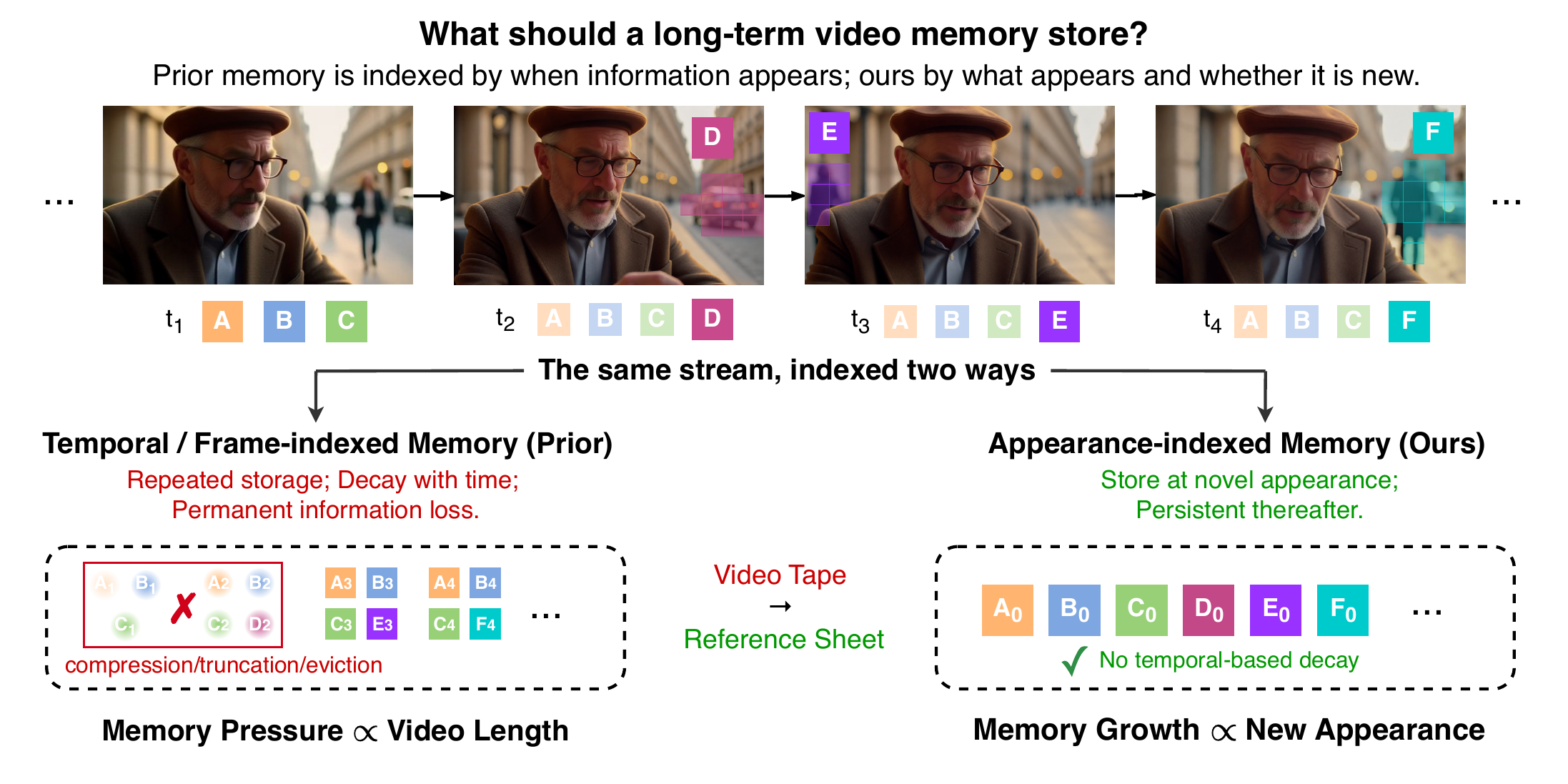}
\caption{\textbf{What should a long-term video memory store?} The same rollout, indexed two ways. \textbf{Top}: the man is recorded at his first appearance ($t_1$); each later frame depicts something new: a passing car ($t_2$), a pedestrian ($t_3$), another passer-by ($t_4$). \textbf{Bottom Left}: frame-indexed memory re-stores old and new content alike at every step, then compresses or evicts it as time passes, so memory pressure grows with video length. \textbf{Bottom Right}: appearance-indexed memory (\ourmodel) stores content when it newly appears, at canonical temporal position $0$, memory scales with the amount of newly appearing content, not with its duration.}
\vspace{-1em}
\label{fig:motivation}
\end{figure*}

\vspace{-2em}
\section{Introduction}
\label{sec:intro}

Generating long videos autoregressively is fundamentally a memory problem. A causal video model produces a video block by block~\cite{diffusionforcing,causvid,selfforcing,causalforcing}, but its attention budget prevents it from revisiting every previously generated frame. At each step, it must decide what to retain and what to discard or compress~\cite{selfforcing,longlive,framepack}. Therefore, we ask a basic question in this paper: \textbf{what should a long video remember, and how should that memory be organized?}

Existing methods organize memory primarily along the axis of \textbf{time} (\cref{fig:motivation}, bottom left). Sliding-window methods retain recent frames and discard older ones, while recency-based caches preserve the near past at high fidelity and allocate progressively less capacity to the distant past~\cite{streamingt2v,framepack,streamingllm} via compression. Although intuitive, this strategy creates a fundamental tension. As earlier content is progressively discarded or compressed, the model loses detailed access to the original visual evidence. Small deviations can therefore accumulate, causing the generated content to gradually drift away from its earlier appearance. Permanently retaining references can mitigate this problem, but excessive reliance on a fixed anchor pulls later generations toward the same appearance and suppresses motion, resulting in freeze. Recency-based memory is therefore caught between forgetting the past and over-anchoring to it.

We instead organize memory around \textbf{appearance novelty} (\cref{fig:motivation}, bottom right). Time-indexed memory asks \textit{when} content was seen; appearance-indexed memory asks \textit{what} was seen that is worth remembering. A long video is not merely a sequence of frames to be compressed according to age; it is a growing cast of subjects, objects, and backgrounds whose identities should remain consistent over time. Memory should therefore preserve content at the moments it becomes newly visible, rather than retaining only its most recent observations: in \cref{fig:motivation}, the man is stored at his first appearance and, while he remains on screen, no already-seen part of him is stored again; newly content (passing car and pedestrian) is admitted the moment it enters. At each \emph{novel appearance event}---content entering the frame, a newly revealed region, or a new viewpoint of a subject already present---we store its KV cache and retain them by novelty rather than by recency. Records compete on how novel they were at admission, never on how recently they were seen. Memory thus grows with the amount of new content, not with video length, and content introduced at second three remains directly accessible at minute five. Because its original keys and values are retained persistently rather than temporally compressed, the model can faithfully reach that content and keep consistent. Long-range consistency thus becomes a consequence of memory organization, rather than something the model must reconstruct.

This principle unifies two mechanisms that may initially appear unrelated. Prior work treats the attention sink (the first few latent frames) as a mechanism for maintaining content consistency~\cite{longlive, causalforcing, selfforcingpp}. We find, however, that its default influence is often insufficient to prevent object and subject drift, tending instead to preserve coarse appearance attributes such as color tone and overall visual style. A reinforced sink is required to maintain subject and object consistency, turning the initial frames into the video's first appearance memory. This interpretation naturally motivates a mechanism for remembering content introduced later in the video. We therefore introduce an optical-flow-based novelty bank, where optical-flow cues~\cite{raft} detect novel appearance events: such as entrances, disocclusions, and newly revealed regions, and store the corresponding KV cache in the memory. The memory unit is a patch at the moment it first appears, rather than an explicitly identified entity. Together, the attention sink and novelty bank instantiate the same principle at different stages of generation: record what appears and retain it.

The two components together alleviate the drift--freeze trade-off. The reinforced sink anchors the opening scene and improves long-range subject and object consistency. The novelty bank continually admits newly introduced content, allowing the model to preserve the consistency of content introduced later, thereby maintaining dynamics. Equipped with these two forms of mechanisms, appearance-indexed memory preserves what has already been established while continuing to admit what is new.

Our main contributions are summarized as follows:

\begin{itemize}
\item \textbf{A new axis for long-video memory: appearance novelty rather than recency.}
We propose organizing autoregressive video memory by appearance novelty instead of temporal recency. Content is recorded at its novel appearance events---its entrance and subsequent novel observations, such as new viewpoints or newly revealed regions---so that each subject accumulates a sparse set of appearance anchors. This formulation alleviates the tension between long-range identity drift and motion freeze.

\item \textbf{\ourmodel, a training-free mechanism.}
\ourmodel combines a permanent attention sink, reinterpreted as memory of the opening scene, with an optical-flow-based novelty bank that extends appearance-indexed memory to new content introduced later in the video.

\item \textbf{Broad validation across models and memory strategies.}
We demonstrate that \ourmodel consistently improves multiple strong video generation baselines~\cite{selfforcing,infiniteforcing,longlive,helios} and outperforms existing temporal recency based memory methods~\cite{deepforcing,memrope,rollingsink}.
\end{itemize}

\section{Related Work}
\label{sec:related}

\noindent\textbf{Autoregressive Video Diffusion Models.}
Latent diffusion~\cite{ldm} with transformer backbones~\cite{dit} has become the standard recipe for high-quality video generation~\cite{vdm,svd,cogvideox,hunyuanvideo,wan21}, with recent systems scaling this paradigm further~\cite{sora}. While most are \emph{bidirectional} and jointly denoise fixed-length clips, recent work enables autoregressive generation by conditioning frames or blocks on past context, often with step distillation for real-time inference~\cite{dmd,dmd2}. Diffusion Forcing~\cite{diffusionforcing} unifies next-token prediction and diffusion through per-token noise levels. CausVid~\cite{causvid} distills a bidirectional teacher into a causal student, while Self-Forcing~\cite{selfforcing} reduces the train--test exposure gap~\cite{scheduledsampling} using student rollouts. Follow-ups extend rollout horizons~\cite{selfforcingpp}, supervise information written into the KV cache~\cite{selfgradientforcing}, and perturb conditioning context for robustness~\cite{incontextforcing}. Other variants improve causal distillation for quality and efficiency~\cite{causalforcing,causalforcingpp,msforcing,opsdv}, physical plausibility~\cite{phys}, or flexible causal--bidirectional generation~\cite{flexforcing}; rolling and chunk-wise methods similarly generate through bounded windows~\cite{rollingdiffusion,magi1,rollingforcing,infiniteforcing}. These methods face a fundamental drift--stability trade-off: persistent early context suppresses drift but can also suppress motion. \ourmodel is complementary to these training-based approaches: it operates entirely at inference on distilled causal models and directly rebalances this trade-off.

\noindent\textbf{Memory for Long Video Generation.}
Long-horizon generation requires preserving information beyond the active attention window. Existing methods organize memory primarily by recency. Temporal/recency-based methods retain recent context through sliding windows or KV caches~\cite{selfforcing,longlive}, introduce short- and long-term anchors~\cite{streamingt2v}, compress past frames temporally~\cite{framepack}, or use chunked continuation for minute-long or nominally infinite generation~\cite{skyreelsv2,skyreelsv3,longcatvideo,ltxvideo,helios}. Relevance-based methods retrieve frames by geometric overlap~\cite{contextasmemory,anchorweave}, route queries to selected context chunks~\cite{mixtureofcontexts}, or maintain adaptive memory banks~\cite{memflow,hydra,wonder}. Related world models preserve long-term geometry and world state~\cite{worldplay,alayaworld,viskoorbis,revisitconsistency,worldtrace}, supported by dedicated corpora~\cite{sekai2}, while other approaches retrain generators with geometric, reference, or reward objectives~\cite{geometryforcing,refalign,vggrpo,sagegrpo,worldcycle}. Multi-shot methods instead maintain consistency across cuts through keyframe or long-context conditioning~\cite{moviedreamer,storyanchors,captaincinema,lct}, cross-shot attention~\cite{holocine,shotadapter,vgot}, explicit shot or entity memory~\cite{storymem,onestory,logishot,videomemory}, positional or streaming modifications~\cite{multishotmaster,shotstream}, and training-free reference or agentic control~\cite{cineweaver,infinitystory,stage}.

A closely related line organizes the KV cache as memory directly. Inspired by attention sinks in streaming language models~\cite{streamingllm} and heavy-hitter eviction~\cite{h2o}, training-free video methods replay the initial sink~\cite{deepforcing,rollingsink}, compress evicted states~\cite{memrope}, gate cache admission by relevance~\cite{tethercache}, or summarize frames by surprise~\cite{surpriseforcing}. SlotMemory~\cite{slotmem} learns object-centric semantic slots as routing addresses for original KV tokens, with prompt- and visual-relevance-based retrieval and eviction. Others restructure caches across heads~\cite{headcast}, absorb history into model weights~\cite{ispa}, modify temporal positional encoding~\cite{lol,infinityrope}, or intervene through spectral anchoring~\cite{freqforcing} and sampling-time guidance~\cite{diffvf,vorchdirector}. Trained counterparts learn related sink-plus-memory structures~\cite{contextforcing,packforcing,relaxforcing}, while Reward Forcing~\cite{rewardforcing} combines compressed memory with motion.

Across these approaches, memory is selected or organized through compression, pruning, semantic structure, or query-dependent relevance. In contrast, we retain original mid-video KV cache according to appearance novelty. Newly appearing content is stored directly, yielding a query-agnostic, novelty-indexed memory bank.
\section{Method}
\label{sec:method}

\ourmodel answers the question: what should a long-term memory remember? First, we revisit causal autoregressive video generation models and their bounded memory (\cref{sec:prelim}). Then, we argue that long-term memory should be indexed by appearance novelty rather than by time, and that content needs be recorded at its novel appearance events (\cref{sec:problem}). Two mechanisms follow: the opening frame is entirely new by construction, so we turn the model's existing attention sink into an explicit \emph{first-impression memory} (\cref{sec:sink}), while every later frame contributes what is new relative to what has been seen, which optical flow detects and a memory bank retains (\cref{sec:bank}). 

\subsection{Preliminaries}
\label{sec:prelim}

We revisit causal, block-wise video generation models such as Self-Forcing~\cite{selfforcing}, which distill a bidirectional video diffusion transformer~\cite{wan21,dit} into a few-step causal generator with distribution matching distillation~\cite{dmd,dmd2}. A video is produced as a stream of blocks of $b$ frames; each frame is patchified into $f$ tokens, and a block is denoised conditioned only on a \emph{bounded, sliding} KV cache of the most recent $L$ frames, implemented as rotary~\cite{rope} self-attention with a fixed-width cache. The bound keeps per-block compute constant and enables arbitrary-length generation.

To counter the resulting drift, sink-augmented backbones such as LongLive~\cite{longlive} keep the first $s$ frames as a permanent attention sink: its tokens stay cached, and their rotary phase is re-applied at retrieval so the anchor never appears at a distance the distilled model never trained on. The sink mechanism itself predates \ourmodel: on backbones that lack it (e.g., vanilla Self-Forcing), our plug-in instantiates the same sink at inference, under the single setting shared across all experiments. Every query attends over
\begin{equation}
  \big[\ \underbrace{\text{Sink}}_{s\text{ frames}}\ \big|\ \underbrace{\text{Prior}}_{L-s-b\text{ frames}}\ \big|\ \underbrace{\text{Current}}_{b\text{ frames}}\ \big].
  \label{eq:layout_base}
\end{equation}

\subsection{What Should a Long Video Remember?}
\label{sec:problem}

The cache in \cref{eq:layout_base} forgets by construction: once a token scrolls past the window, it is gone, and the opening frame is its sole survivor. As more of the early visual context is discarded, the model retains progressively less information about the original appearance. Small inconsistencies can therefore accumulate over time, gradually causing the generated content to drift from its earlier identity.

The prevailing diagnosis is recency-based memory. Sliding windows keep the most recent frames and discard the rest; recency caches and importance-by-recency compression keep the near past at full fidelity and squeeze the far past into a shrinking budget~\cite{streamingt2v,framepack,streamingllm}. Recent training-free methods follow the same axis: widening the sink and pruning by attention participation~\cite{deepforcing}, compressing evicted keys into a few memory tokens~\cite{memrope}, or cyclically replaying the states of the first seconds~\cite{rollingsink}. They share one assumption: that rendering a subject consistently at $t{=}60$\,s means reaching back through the timeline to when it was last seen, so the remedy
is to store more of that timeline. 

We argue the premise is wrong. A long video does not need its timeline; it needs its \emph{cast}. An entity's appearance is established by a sparse set of \emph{novel appearance events}---its entry and the later sightings of newly revealed parts---and nothing about the intervening forty seconds is required to render it faithfully to keep consistency. Memory should therefore be indexed by appearance: record content when it is newly revealed, and allocate the fixed capacity by novelty rather than by recency, so that what survives is decided by what appeared, not by when it appeared.

This raises a natural question: what counts as \emph{new}? The answer has two exhaustive cases. At $t{=}0$ nothing has been seen, so the opening frame is new in its entirety (\cref{sec:sink}); afterwards, content is new only \emph{relative to what has already appeared}---an object entering the frame, a region disoccluded by motion, a style introduced by a change of scene (\cref{sec:bank}).

\subsubsection{The First Frame: from Stability Anchor to First-Impression Memory}
\label{sec:sink}

The opening frame is entirely novel content, and sink-augmented backbones already retain it permanently as the attention sink; where the backbone has none, the plug-in instantiates one (\cref{sec:prelim}). Attention sinks are known to stabilize streaming generation, which motivates their adoption~\cite{streamingllm,longlive}. But it is worth asking what they actually stabilize. We find that, under the default attention distribution, later queries assign insufficient weight to the sink. As a result, the sink mainly anchors global appearance characteristics (color tone, background, and overall style )and helps prevent quality degradation, while the identity of subjects introduced in the opening scene can still drift. 

We therefore strengthen the sink attention, promoting it from a stability anchor to a persistent \emph{first-impression memory} that is more strongly attended to by queries, thereby better preserving subject identity over time. For a query $\mathbf{q}$ attending to keys $\{\mathbf{k}_j\}$, we add a fixed positive bias to the pre-softmax logits:
\begin{equation}
  \alpha_j = \operatorname{softmax}_j\!\left(
  \frac{\mathbf{q}^\top \mathbf{k}_j}{\sqrt{d}} + \beta_j \right),
  \beta_j =
  \begin{cases}
    \ln \lambda, & j \in \mathrm{Sink}, \\
    0, & \text{otherwise},
  \end{cases}
  \label{eq:sink}
\end{equation}
where $\lambda>1$ is the reinforcement factor. Adding $\ln\lambda$ multiplies each sink key's unnormalized attention weight by $\lambda$, uniformly increasing the sink's prior importance while preserving content-based selection through query--key similarity. Applying the bias before the softmax keeps the distribution normalized: sink keys still compete with all other keys, and attention elsewhere is compressed proportionally. The bias is also directly compatible with fused attention kernels. Its magnitude is a measurement rather than a choice: we set $\lambda=5$, with its effect analyzed later in \cref{fig:lambda_sweep}.

\subsubsection{Every Frame After: An Optical-Flow Novelty Bank}
\label{sec:bank}

The reinforced sink covers the opening scene, but a long video continually introduces new content: a second subject enters the frame, a camera pan reveals a storefront, or an occluded region becomes visible. After $t{=}0$, ``new'' therefore means content that has not been observed before, making novelty detection a property of the video stream rather than of the model's current query. Optical flow provides a direct cue: a patch is novel when it enters the field of view, becomes disoccluded without a traceable antecedent, or reveals previously unseen content under a change in viewpoint. We detect these appearance events with RAFT~\cite{raft} and retain the most novel patches observed so far in a fixed set of $K$ additional KV slots. Admission is query-agnostic: patches are retained according to visual appearance novelty rather than query--key affinity, without entity grouping or re-identification. This forms the memory bank: 
\begin{equation}
  \big[\ \text{Sink}\ \big|\ \underbrace{\text{Memory Bank}}_{K\text{ slots}}\ \big|\ \text{Prior}\ \big|\ \text{Current}\ \big].
  \label{eq:layout_bank}
\end{equation}
The bank capacity $K$ is a configurable memory budget rather than an architectural constant. For the one-minute videos considered in our experiments, we find that one frame's worth of tokens ($K{=}1{,}560$) provides a good trade-off and use this setting throughout; see \cref{tab:bank_sink_sweeps} for the ablation. Longer videos, or videos with more frequent novel appearances, can naturally allocate additional slots to accommodate the increased memory demand.

\begin{figure}[t]
  \centering
  \includegraphics[width=\linewidth, trim={2em 2em 2em 2em}]{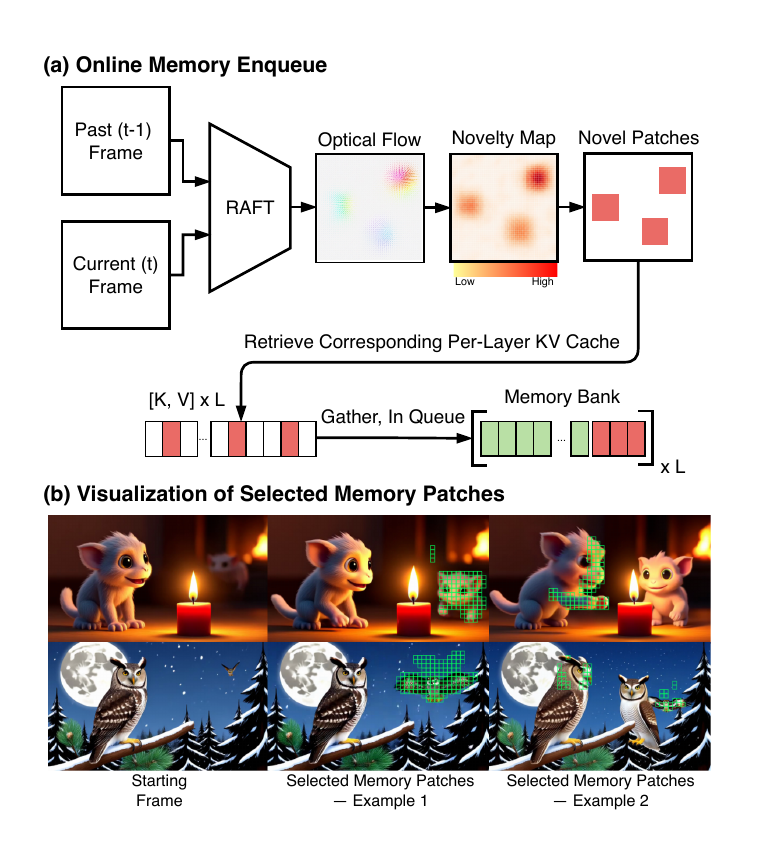}
  \caption{\textbf{The optical-flow novelty bank.} \textbf{(a)} For every pair of consecutive frames, RAFT
  flow is aggregated into the per-patch novelty score of \cref{eq:novelty}, and the newly appearing
  patches are selected. Their per-layer key/value vectors are gathered from the KV cache (temporal RoPE
  phase reset to zero) and added to the bank together with their scores; the bank keeps the top-$K$
  ($K{=}1{,}560$) by score. \textbf{(b)} Selected memory patches (green) on two examples: entering
  subjects and newly visible parts of subjects.}
  \label{fig:method}
\end{figure}

\noindent\textbf{Appearance Novelty.} For each newly decoded frame $x_t$, we estimate forward and backward RAFT flow to its predecessor $x_{t-1}$, where $F_{a\to b}(u)$ denotes the optical-flow displacement from pixel $u$ in frame $a$ to frame $b$. A pixel $u$ of $x_t$ is regarded as newly appeared when its backtrace $u' = u + F_{t\to t-1}(u)$ finds no reliable antecedent in $x_{t-1}$: the forward--backward cycle residual $e_{\mathrm{cyc}}(u)=\lVert F_{t\to t-1}(u)+F_{t-1\to t}(u')\rVert$ exceeds $\tau_{\mathrm{cyc}}$, the photometric residual $e_{\mathrm{pho}}(u)=\lvert x_t(u)-x_{t-1}(u')\rvert$ exceeds $\tau_{\mathrm{pho}}$, or $u'$ falls outside the frame. Such pixels receive
\begin{equation}
\nu(u)=e_{\mathrm{cyc}}(u)+\alpha e_{\mathrm{pho}}(u),
\label{eq:novelty}
\end{equation}
where $\alpha$ balances the two residuals; all other pixels score $0$.

The map of already-claimed novelty is propagated with the flow, so continuously traceable content fires only once, when it first appears. If a region returns after its flow correspondence has been broken, it can register as a new appearance event and receive a new anchor. For a patch $p$,
\begin{equation}
s(p)=\max_{u\in p}\nu(u),\qquad
\hat{s}(p)=\frac{s(p)}{\max(\rho_b,1)},
\end{equation}
where $\rho_b$ is the $90^{\text{th}}$ percentile of positive patch scores in temporal block $b$. We use $\tau_{\mathrm{cyc}}=2$ px, $\tau_{\mathrm{pho}}=0.2$, and $\alpha=0.5$ throughout. This block-wise normalization prevents brief bursts of large motion from dominating the fixed budget, so admissions depend on relative novelty rather than absolute motion magnitude.

As illustrated in \cref{fig:method} (a), each newly decoded frame is compared with its predecessor using RAFT to obtain a novelty map, from which newly appearing patches are selected. For each selected patch, we copy the corresponding key/value vectors from every attention layer of the KV cache and store them with their frozen novelty scores. At retrieval, bank keys use temporal rotary phase $0$ while retaining their original spatial phase, avoiding unseen long-range temporal RoPE offsets and treating the bank as a time-neutral appearance memory. The bank is then updated with a simple top-$K$ rule: among all candidates observed so far, only the $K$ highest-scoring patches are retained. Thus, memory retention is time-neutral and parameter-free: a patch remains in the bank according to its appearance novelty rather than how recently it was observed, without requiring a learned controller or an explicit recency term.

\begin{table*}[!th]
\centering
\scriptsize
\setlength{\tabcolsep}{2.6pt}
\caption{
\textbf{Main results on VBench-Long~\cite{vbenchlong}, all 16 dimensions} ($128$ prompts $\times\,5$ seeds, $60$\,s; higher is better). \ourmodel (optical-flow memory bank $+$ reinforced sink) is applied as the same training-free inference hook to every base model, each followed by its ``$+$ Ours'' row; for Self-Forcing we additionally report a plain, unreinforced sink ($\lambda{=}1$).
Helios is 14B, all others are 1.3B.
Abbreviations (first letter of each word): S.C.\ Subject Consistency, B.C.\ Background Consistency, T.F.\ Temporal Flickering, M.S.\ Motion Smoothness, D.D.\ Dynamic Degree, A.Q.\ Aesthetic Quality, I.Q.\ Imaging Quality; semantic suite: O.C.\ Object Class, M.O.\ Multiple Objects, H.A.\ Human Action, C.\ Color, S.R.\ Spatial Relationship, S.\ Scene, A.S.\ Appearance Style, T.S.\ Temporal Style, Ov.C.\ Overall Consistency. Q./Sem./Total $=$ official VBench aggregation (min--max normalized; quality $=7$ dims with Dynamic Degree half-weighted, semantic $=9$ dims, Total $=(4\,\mathrm{Q}+\mathrm{Sem})/5$).
All values are percentages.
}
\label{tab:vbench_long_main}
\resizebox{\textwidth}{!}{
\begin{tabular}{lccccccc|ccccccccc|ccc}
\toprule
\textbf{Method} & \textbf{S.C.} & \textbf{B.C.} & \textbf{T.F.} & \textbf{M.S.} & \textbf{D.D.} &
\textbf{A.Q.} & \textbf{I.Q.} & \textbf{O.C.} & \textbf{M.O.} & \textbf{H.A.} & \textbf{C.} &
\textbf{S.R.} & \textbf{S.} & \textbf{A.S.} & \textbf{T.S.} & \textbf{Ov.C.} &
\textbf{Q.} & \textbf{Sem.} & \textbf{Total} \\
\midrule
Self-Forcing~\cite{selfforcing}  & 98.0 & 96.8 & 98.3 & 98.9 & 27.5 & 53.8 & 69.2 & 79.5 & 32.1 & 77.0 & 72.8 & 62.6 & 12.1 & 20.6 & 20.9 & 19.6 & 80.4 & 58.0 & 75.9  \\
\quad $+$ Sink ($\lambda{=}1$) & 98.2 & 97.0 & 98.3 & 98.9 & 30.4 & 53.3 & 67.9 & 89.2 & 41.1 & 75.1 & 78.2 & 73.6 & 14.6 & 20.4 & 20.3 & 21.2 & 80.4 & 62.3 & 76.8  \\
\quad $+$ \textbf{Ours} & 97.7 & 96.3 & 97.4 & 98.5 & 58.1 & 58.4 & 70.9 & 91.8 & 47.3 & 77.1 & 94.6 & 68.4 & 17.7 & 18.7 & 22.8 & 22.1 & 82.9 & 65.5 & \textbf{79.5}  \\
\midrule
Infinite-Forcing~\cite{infiniteforcing} & 97.4 & 95.9 & 97.1 & 98.4 & 61.3 & 55.8 & 68.8 & 93.6 & 49.9 & 74.1 & 81.9 & 64.0 & 12.3 & 18.6 & 22.9 & 21.9 & 82.2 & 62.9 & 78.3  \\
\quad $+$ \textbf{Ours} & 97.4 & 96.0 & 97.2 & 98.4 & 71.3 & 57.5 & 68.8 & 97.1 & 48.9 & 75.7 & 96.1 & 63.4 & 15.1 & 18.5 & 23.0 & 22.2 & 83.3 & 65.4 & \textbf{79.7}  \\
\midrule
LongLive~\cite{longlive} & 98.7 & 97.0 & 98.5 & 99.2 & 20.3 & 57.1 & 71.6 & 98.2 & 52.2 & 75.9 & 80.4 & 80.5 & 14.8 & 18.1 & 22.7 & 22.6 & 81.2 & 65.9 & 78.1  \\
\quad $+$ \textbf{Ours} & 98.5 & 97.0 & 98.5 & 99.1 & 26.6 & 57.3 & 71.0 & 98.4 & 50.7 & 76.7 & 82.7 & 80.4 & 19.5 & 17.9 & 23.0 & 22.8 & 81.5 & 66.8 & \textbf{78.5}  \\
\midrule
Helios~\cite{helios} & 95.8 & 97.4 & 98.6 & 99.1 & 36.2 & 45.2 & 49.0 & 36.1 & 8.9 & 66.7 & 53.6 & 49.0 & 4.6 & 21.2 & 20.3 & 21.3 & 76.6 & 45.4 & 70.3  \\
\quad $+$ \textbf{Ours} & 96.4 & 97.3 & 98.1 & 98.8 & 47.8 & 53.6 & 60.0 & 83.7 & 28.5 & 80.0 & 72.2 & 63.4 & 18.6 & 19.3 & 22.2 & 23.7 & 80.2 & 60.5 & \textbf{76.3}  \\
\bottomrule
\end{tabular}
}
\vspace{-1em}
\end{table*}

\begin{table*}[!th]
\centering
\scriptsize
\setlength{\tabcolsep}{2.6pt}
\caption{
\textbf{Comparison of \ourmodel with other training-free methods.} Same evaluation settings on VBench-Long as in \cref{tab:vbench_long_main}.
}
\label{tab:vbench_long_plugins}
\resizebox{\textwidth}{!}{
\begin{tabular}{lccccccc|ccccccccc|ccc}
\toprule
\textbf{Method} & \textbf{S.C.} & \textbf{B.C.} & \textbf{T.F.} & \textbf{M.S.} & \textbf{D.D.} &
\textbf{A.Q.} & \textbf{I.Q.} & \textbf{O.C.} & \textbf{M.O.} & \textbf{H.A.} & \textbf{C.} &
\textbf{S.R.} & \textbf{S.} & \textbf{A.S.} & \textbf{T.S.} & \textbf{Ov.C.} &
\textbf{Q.} & \textbf{Sem.} & \textbf{Total} \\
\midrule
Self-Forcing & 98.0 & {96.8} & {98.3} & 98.9 & 27.5 & 53.8 & 69.2 & 79.5 & 32.1 & 77.0 & 72.8 & 62.6 & 12.1 & {20.6} & 20.9 & 19.6 & 80.4 & 58.0 & 75.9  \\
$+$ Deep Forcing~\cite{deepforcing} & {98.4} & 93.7 & 98.2 & 94.8 & 32.0 & 55.9 & 71.8 & {98.3} & {48.5} & 76.6 & 82.9 & {73.1} & {21.8} & 18.3 & {23.0} & {22.3} & 78.7 & {66.1} & 76.2  \\
$+$ MemRoPE~\cite{memrope} & {98.4} & 96.7 & 98.0 & {99.0} & {41.2} & 56.6 & {73.1} & 94.2 & 44.1 & 76.6 & 84.8 & 68.4 & 15.9 & {19.1} & 21.8 & 21.1 & {82.4} & 63.6 & 78.7  \\
$+$ Rolling Sink~\cite{rollingsink} & {98.9} & {97.3} & {98.7} & {99.2} & 23.0 & {58.3} & {72.6} & {97.9} & {48.2} & {80.7} & {90.3} & {76.5} & 17.6 & 18.2 & {23.2} & {22.5} & 81.8 & 67.1 & {78.9}  \\
$+$ {Ours} & 97.7 & 96.3 & 97.4 & 98.5 & 58.1 & {58.4} & 70.9 & 91.8 & 47.3 & {77.1} & {94.6} & 68.4 & {17.7} & 18.7 & 22.8 & 22.1 & {82.9} & 65.5 & \textbf{79.5}  \\
\bottomrule
\end{tabular}
}
\vspace{-1em}
\end{table*}

\cref{fig:method} (b) illustrates what the criterion admits in practice. Slots are assigned to novel appearances rather than motion magnitude or recency. A newly entering subject is captured as soon as it becomes visible, as with the second monster and the additional owls. The same rule also admits newly revealed parts of an existing subject, such as the first monster's head and paws or the first owl's turned face, preserving appearance cues that a recency window would discard after the pose changes. Continuously traceable, unchanged regions do not repeatedly trigger the detector, so the fixed budget is reserved for newly revealed appearance cues that later frames may need to match.

\section{Experiments}
\label{sec:exp}

\newcommand{\NA}{--}

\noindent\textbf{Setup.} We evaluate on VBench-Long~\cite{vbench,vbenchlong} and report its sixteen long-video metrics. In order to obtain a stable comparison, we average over $128$ prompts and $5$ different seeds. Videos are $60$ seconds ($240$ latent frames).

\subsection{Quantitative Results}

\noindent\textbf{Main Results on VBench-Long.}
In \cref{tab:vbench_long_main}, we report \ourmodel's performance based on four different video generation models: Self-Forcing~\cite{selfforcing}, Infinite-Forcing~\cite{infiniteforcing}, LongLive~\cite{longlive}, and Helios~\cite{helios}. On the Self-Forcing baseline, adding a plain, unreinforced sink already helps ($75.9\to76.8$ Total) but recovers little motion (Dynamic Degree $27.5\to30.4$); \ourmodel more than doubles the dynamic degree to $58.1$ while improving aesthetic and imaging quality. On Infinite Forcing~\cite{infiniteforcing}, whose sink already raises dynamic degree to $61.3$, \ourmodel further increases it to $71.3$ ($+10.0$), with a clear gain in overall score from $78.3$ to $79.7$. Compared with the $1.3$B LongLive~\cite{longlive} and the $14$B Helios~\cite{helios}, \ourmodel can still consistently improve the performance. Thus, \ourmodel improves long-video dynamics on both backbones without sacrificing consistency or visual quality. 

\noindent\textbf{Comparison with Other Training-Free Methods.}
We compare \ourmodel with three latest training-free long video generation methods on Self-Forcing in \cref{tab:vbench_long_plugins}. While Deep Forcing~\cite{deepforcing}, MemRoPE~\cite{memrope}, and Rolling Sink~\cite{rollingsink} improve selected metrics, they recover substantially less motion, with Dynamic Degree ranging from $23.0$ to $41.2$ versus $58.1$ for \ourmodel. \ourmodel achieves the best overall score ($79.5$) while remaining competitive in temporal consistency. This suggests that \ourmodel provides a better balance of long-term dynamics, stability, and visual quality.

\subsection{Ablation Study}

We use Infinite-Forcing~\cite{infiniteforcing} for the ablation study because vanilla Self-Forcing~\cite{selfforcing} has no sink frame: attaching \ourmodel to it \emph{instantiates} the sink, so a component ablation there would conflate creating the sink with reinforcing it. Infinite-Forcing, a Self-Forcing variant retrained with a sink frame, already carries the sink, so the reinforcement and the bank can be ablated cleanly.

\vspace{1em}
\noindent\textbf{Ablation of Components.}
\cref{tab:vbench_long_ablation} reveals a clear temporal division of labor between the two components. The reinforced sink mainly anchors information established early in the generation, which improves consistency and semantic quality, but can overly constrain later evolution, as reflected by the drop in Dynamic Degree ($61.3\to41.6$). In contrast, the memory bank helps retain and incorporate information that emerges or evolves over time, substantially increasing Dynamic Degree ($61.3\to78.2$), with only minor reductions in consistency and visual quality. When used together, the reinforced sink provides a stable foundation for earlier content, while the memory bank allows later and more dynamic content to develop without drifting from it. As a result, the full model achieves the best overall score ($79.7$).

\begin{table}[!th]
\centering
\scriptsize
\setlength{\tabcolsep}{2.6pt}
\caption{
{\textbf{Ablation of the reinforced sink and memory bank.}}
}
\label{tab:vbench_long_ablation}

\resizebox{\columnwidth}{!}{
\begin{tabular}{cccccccc|ccc}
\toprule

\makecell[c]{{\textbf{Reinf.}}\\{\textbf{Sink}}}
&
\makecell[c]{{\textbf{Memory}}\\{\textbf{Bank}}}
&
\makecell[c]{{\textbf{S.C.}}}
&
\makecell[c]{{\textbf{B.C.}}}
&
\makecell[c]{{\textbf{M.S.}}}
&
\makecell[c]{{\textbf{D.D.}}}
&
\makecell[c]{{\textbf{A.Q.}}}
&
\makecell[c]{{\textbf{I.Q.}}}
&
\makecell[c]{{\textbf{Q.}}}
&
\makecell[c]{{\textbf{Sem.}}}
&
\makecell[c]{{\textbf{Total}}}
\\

\midrule

\xmark & \xmark
& {97.4}
& 95.9
& {98.4}
& 61.3
& 55.8
& {68.8}
& 82.2
& 62.9
& 78.3
\\

\cmark & \xmark
& {98.2}
& {96.7}
& {98.8}
& 41.6
& {57.8}
& {68.9}
& 81.9
& {65.2}
& 78.6
\\

\xmark & \cmark
& 96.8
& 95.5
& 98.1
& {78.2}
& 55.2
& 68.5
& {82.8}
& 63.0
& {78.8}
\\

\cmark & \cmark
& {97.4}
& {96.0}
& {98.4}
& {71.3}
& {57.5}
& {68.8}
& {83.3}
& {65.4}
& \textbf{79.7}
\\

\bottomrule
\end{tabular}
}
\end{table}

\begin{figure}[!ht]
  \centering
  \includegraphics[width=0.95\linewidth, trim={1em 0em 1em 0em}]{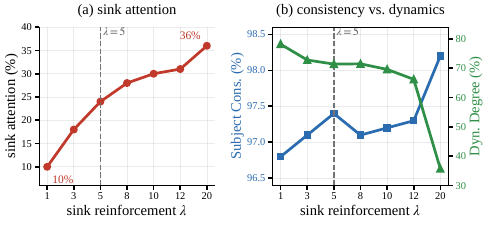}
  \caption{\textbf{$\lambda$ Sweep} (on $128$ prompts $\times\,5$ seeds). \textbf{(a)} $\lambda$ smoothly sets attention on the sink: ${\sim}24\%$ at $\lambda{=}5$. \textbf{(b)} Subject Consistency is already saturated while Dynamic Degree erodes past $\lambda{\approx}8$.}
  \vspace{-1em}
  \label{fig:lambda_sweep}
\end{figure}

\noindent\textbf{Ablation on Bank Capacity and Sink-Reinforcement Strength.}
As shown in~\cref{tab:bank_sink_sweeps}, increasing the bank capacity from $f=1$ to $2$ or $4$ yields only marginal improvements in the Total score, while a larger capacity of $f=8$ reduces it to $79.3$. Given the additional memory and computational overhead of larger banks, we use $f=1$ by default.
For sink-reinforcement strength, increasing $\lambda$ from $1$ to $5$ improves the Total score from $78.8$ to $79.7$, whereas further increasing $\lambda$ consistently degrades performance, yielding $79.2$, $78.9$, and $78.6$ at $\lambda=8,10,12$, respectively. We therefore set $\lambda=5$ as the default. The diagnostic sweep in~\cref{fig:lambda_sweep} further explains this trade-off: increasing $\lambda$ progressively directs more attention to the sink, while Subject Consistency remains largely saturated and Dynamic Degree begins to deteriorate under stronger reinforcement. This suggests that moderate sink reinforcement is sufficient to leverage the opening-scene information, whereas excessive reinforcement can over-constrain the model's dynamics. 

\begin{table}[!ht]
\centering
\scriptsize
\setlength{\tabcolsep}{2.6pt}
\caption{
\textbf{Ablations on bank capacity and sink-reinforcement strength.} Bold columns denote the default settings.
}
\label{tab:bank_sink_sweeps}
\resizebox{\linewidth}{!}{
\begin{tabular}{lcccccc|cccccc}
\toprule
& \multicolumn{6}{c}{\textbf{Bank Capacity (frame)}}
& \multicolumn{6}{c}{\textbf{Sink-Reinforcement Strength (\(\lambda\))}} \\
\textbf{Metric}
& 1/4
& 1/2
& \textbf{1}
& 2
& 4
& 8
& 1
& 3
& \textbf{5}
& 8
& 10
& 12 \\
\midrule
\textbf{Q.}
& 83.0 & 83.2 & \textbf{83.1} & 83.4 & 83.3 & 82.9
& 82.8 & 83.0 & \textbf{83.1} & 82.7 & 82.5 & 82.2  \\
\textbf{Sem.}
& 66.0 & 65.8 & \textbf{65.7} & 65.4 & 65.9 & 65.1
& 63.0 & 64.5 & \textbf{65.7} & 65.2 & 64.4 & 64.1  \\
\textbf{Total}
& 79.6 & 79.7 & \textbf{79.7} & 79.8 & 79.8 & 79.3
& 78.8 & 79.3 & \textbf{79.7} & 79.2 & 78.9 & 78.6  \\
\bottomrule
\end{tabular}
}
\end{table}

\subsection{Human Study}

We further conduct a human study to evaluate long video quality beyond automatic metrics.
We have $9$ participants, each of whom completes $24$ blind pairwise comparisons, resulting in $216$ judgments in total. In each trial, participants are shown two anonymized videos side by side. We compare \ourmodel against baseline~\cite{selfforcing} and three competitors~\cite{deepforcing, memrope, rollingsink}. Participants evaluate each pair along three dimensions: \emph{Consistency}, \emph{Motion}, and \emph{Overall} quality. Judgments marked as \emph{HARD TO SAY} are treated as ties when computing win rates, with each tie contributing half a win. Specifically, the win rate is computed as $(W + 0.5T) / N$, where $W$ and $T$ denote the numbers of wins and \emph{HARD TO SAY} judgments, respectively, and $N$ is the total number of comparisons. As shown in~\cref{tab:human_study}, \ourmodel is consistently preferred over all competing methods. These results show that the improvements of \ourmodel are clearly perceptible to human. 

\begin{table}[!ht]
\centering
\caption{\textbf{Human study results.}
Win rates of \ourmodel in blind pairwise comparisons against each competitor. ``T'' denotes the number of \textit{HARD TO SAY} judgments, which are treated as ties and assigned 0.5 credit to win rates.}
\label{tab:human_study}
\resizebox{\linewidth}{!}{
\begin{tabular}{lcccc}
\toprule
\textbf{Competitor} & \textbf{Consistency} & \textbf{Motion} & \textbf{Overall} & $\mathbf{n}$ \\
\midrule
Self-Forcing~\cite{selfforcing} & $85.2\%$ (6T) & $76.9\%$ (7T) & $84.3\%$ (3T) & 54 \\
Deep Forcing~\cite{deepforcing} & $73.1\%$ (15T) & $76.9\%$ (5T) & $77.8\%$ (4T) & 54 \\
MemRoPE~\cite{memrope} & $84.3\%$ (7T) & $67.6\%$ (5T) & $75.0\%$ (3T) & 54 \\
Rolling Sink~\cite{rollingsink} & $82.4\%$ (7T) & $76.9\%$ (5T) & $79.6\%$ (6T) & 54 \\
\midrule
\textbf{All} & $\mathbf{81.3\%}$ (35T) & $\mathbf{74.5\%}$ (22T) & $\mathbf{79.2\%}$ (16T) & \textbf{216} \\
\bottomrule
\end{tabular}
}
\vspace{-0.5em}
\end{table}

\subsection{Qualitative Results}

\begin{figure*}[!th]
  \centering
  \includegraphics[width=\linewidth]{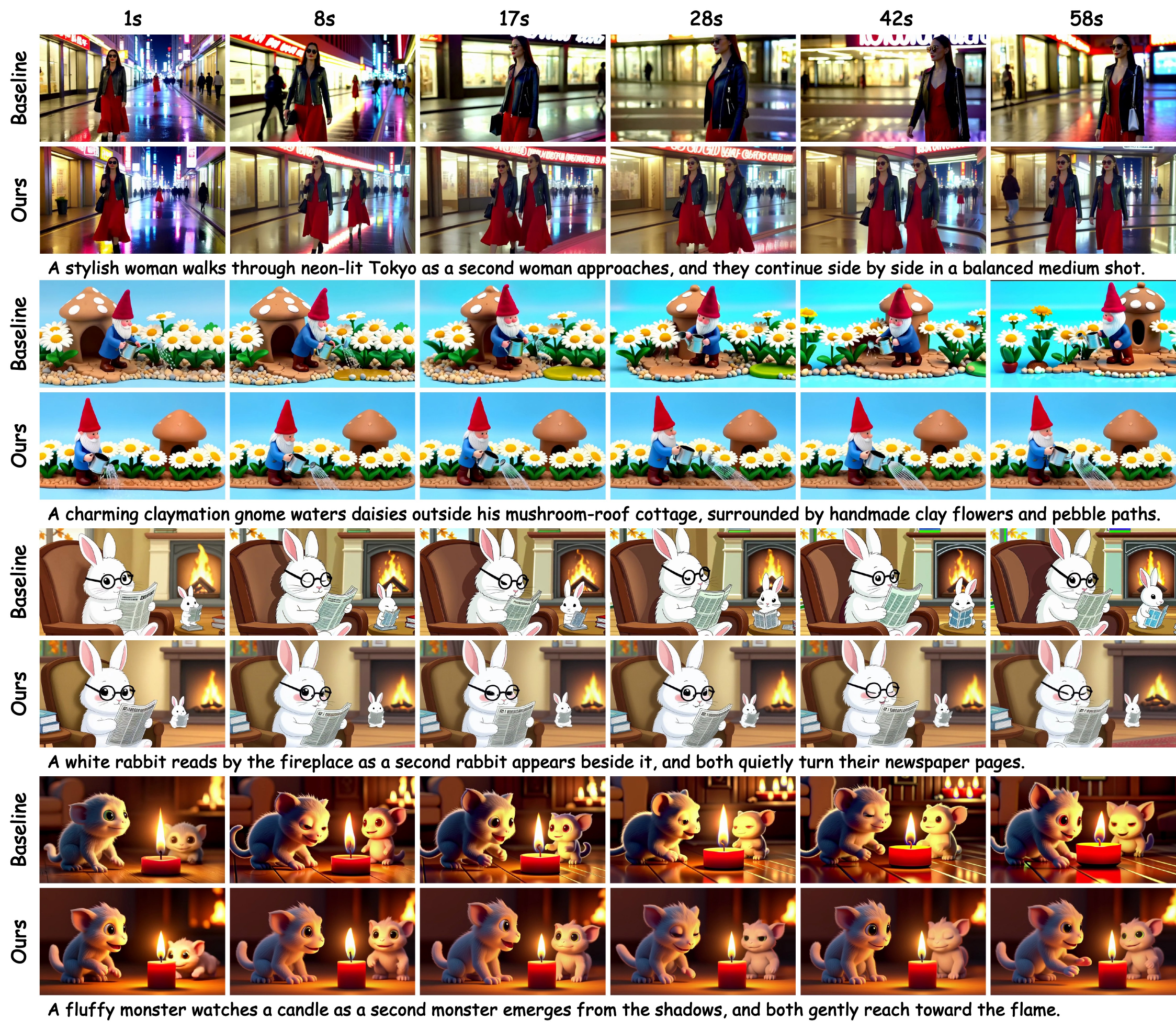}
  \caption{\textbf{Qualitative results} on four one-minute prompts. Across all scenes, the baseline loses subject identity, style, scale, or background consistency. In contrast, \ourmodel preserves the subject(s) and the full scene throughout the minute.}
  \vspace{-1em}
  \label{fig:qual}
\end{figure*}

\cref{fig:qual} compares one-minute rollouts with and without \ourmodel on four prompts spanning photographic, claymation, and cartoon styles. The characteristic failures of recency memory surface by mid-video and compound from there: subjects lose identity or duplicate, a second subject that enters mid-video never stabilizes, and the scene's style and layout gradually migrate away from the prompt. With \ourmodel, the same base model holds subject identity, scene, and style through the full minute, including subjects introduced long after the opening frame, while the motion remains natural rather than frozen. Full videos and additional examples are available on the project page.

\section{Conclusion and Limitations}
\label{sec:conclusion}

We presented \ourmodel, a training-free memory mechanism that organizes finite memory by appearance novelty rather than temporal recency. It combines a reinforced attention sink that preserves the opening scene with an optical-flow-based novelty bank that retains newly appearing content. Applied to frozen causal backbones, \ourmodel improves long-horizon dynamics while maintaining visual quality and consistency, outperforming existing training-free memory methods.

More broadly, our results suggest that appearance novelty provides a useful principle for organizing memory in long-form video generation, while motivating richer notions of novelty beyond the local correspondence used in our current implementation. Longer-range correspondence or higher-level semantic grouping could enable more selective and persistent memory allocation over extended generation. We hope this perspective provides a simple and general foundation for developing more adaptive memory mechanisms for long-form video generation.

Still, it relies on optical-flow correspondence, which may therefore repeatedly mark stochastic textures, such as rain, spray, or rippling water, as novel, consuming memory without adding meaningful new information.

\noindent \textbf{Acknowledgement.} This work is supported by U.S. National Science Foundation Award IIS-2433768 and IIS-2127544. This work also used DeltaAI at National Center for Supercomputing Applications (NCSA) through allocation CIS260420 from the Advanced Cyberinfrastructure Coordination Ecosystem: Services \& Support (ACCESS) program, which is supported by U.S. National Science Foundation grants \#2138259, \#2138286, \#2138307, \#2137603, and \#2138296.

{
    \small
    \bibliographystyle{ieeenat_fullname}
    \bibliography{main}
}

\clearpage
\setcounter{section}{0}
\renewcommand{\thesection}{\Alph{section}}
\maketitlesupplementary

\label{sec:suppl}

\section{Implementation Details}
\label{sec:suppl_impl}

\noindent\textbf{Backbone and Rollout.}
Our primary baseline is Self-Forcing~\cite{selfforcing}, a DMD-distilled~\cite{dmd, dmd2} causal autoregressive video diffusion model built on Wan2.1-T2V-1.3B~\cite{wan21}. The model generates $832{\times}480$ video block by block ($3$ latent frames per block) with a sliding KV cache. Each latent frame is a $30{\times}52$ grid of $1{,}560$ tokens, and the attention window is set to cover $6$ latent frames: one persistent sink frame, two prior frames, and the three-frame current block. A $60$-second video is a rollout of $240$ latent frames ($80$ blocks). The other backbones in \cref{tab:vbench_long_main} use their public checkpoints and inference code unchanged, with \ourmodel attached as the same inference hook.

\noindent\textbf{Reinforced Sink.}
The sink's un-normalized attention weight is multiplied by $\lambda{=}5$, implemented by adding $\log\lambda$ to the sink logits before the softmax.

\noindent\textbf{Memory Bank.}
The bank holds $K{=}1{,}560$ slots. Per-pixel appearance novelty follows \cref{eq:novelty} (RAFT-small~\cite{raft} at half resolution, $12$ refinement iterations), max-pooled onto the $30{\times}52$ patch grid. Slot contents follow the indexed-copy rule: for each chosen position $(t,h,w)$, every attention layer re-inserts the exact key/value that token wrote to the cache, with temporal rotary phase $0$ and the original spatial phase. Refills are incremental: only newly admitted positions are copied, and only activations of still-novel tokens are retained in the history, which keeps the per-block cost constant (\cref{sec:suppl_speed}).

\noindent\textbf{Training-free with One Shared Setting.}
\ourmodel introduces no learned parameters and requires no fine-tuning. We use the same hyperparameter setting (\cref{tab:impl}) across all backbones, video lengths, and prompt sets.

\begin{table}[th]
\centering
\small
\setlength{\tabcolsep}{6pt}
\caption{\textbf{Hyperparameters of \ourmodel}.}
\label{tab:impl}
\begin{tabular}{lc}
\toprule
\textbf{Hyperparameter} & \textbf{Value} \\
\midrule
Sink size & $1$ latent frame \\
Attention window & $6$ latent frames \\
Block size & $3$ latent frames \\
Bank capacity $K$ & $1{,}560$ tokens \\
Sink reinforcement $\lambda$ & $5$ \\
Flow cycle-error threshold $\tau_{\mathrm{cyc}}$ & $2$ px \\
Photometric-error threshold $\tau_{\mathrm{pho}}$ & $0.2$ \\
Photometric weight $\alpha$ (\cref{eq:novelty}) & $0.5$ \\
Score normalization & per-block $90^{\text{th}}$-pct. \\
\bottomrule
\end{tabular}
\end{table}

\section{Length Generalization}
\label{sec:suppl_length}

\begin{figure}[!ht]
\centering
\includegraphics[width=\linewidth, trim={0em 0em 0em 1em}]
{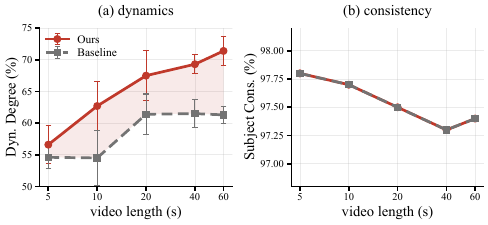}
\caption{\textbf{Length Sweep}
(Baseline vs.\ \ourmodel, $5$--$60$s; error bars show $\pm$std over $5$ seeds).
\textbf{(a)} Our method maintains higher Dynamic Degree across all lengths.
\textbf{(b)} Subject Consistency remains comparable to the baseline.}
\label{fig:length}
\end{figure}

\cref{fig:length} evaluates video durations from $5$ to $60$ seconds. Our method consistently achieves higher Dynamic Degree across all lengths, with the gain reaching $+10.1$ points at $60$s. Meanwhile, Subject Consistency remains comparable to the baseline. This shows that our method sustains strong dynamics in longer videos without sacrificing consistency.

\section{Appearance Memory and Identity Preservation}
\label{sec:exp_concept}

We next examine whether the appearance information stored by \ourmodel remains useful over long horizons. We consider increasingly demanding settings: identity preservation under continuous visibility, retention over substantially extended rollouts, and controlled reappearance after the subject has been absent for much longer than the local attention window.

\subsection{Long-Horizon Subject Identity}
\label{sec:suppl_protocols}

We stage $20$ prompts in which a distinctive subject stays on screen for the full minute ($\times\,5$ seeds per arm) and measure whether it still looks like itself at the end. Identity is measured by DINOv2~\cite{dinov2} \texttt{CLS} similarity between an early window ($2$--$8$\,s) and the final $10$\,s, computed both on the full frame and on the subject crop obtained from a Qwen2.5-VL detector (top-$1$ box for the subject phrase with a $15\%$ margin; detection rate $97\%$). The two arms are compared pairwise per prompt and seed. \ourmodel clearly improves both identity measures (\cref{tab:identity_long}).

\begin{table}[!bth]
\centering
\setlength{\tabcolsep}{5pt}

\caption{
\textbf{Long-horizon Subject Identity} ($20$ prompts $\times\,5$ seeds, $60$\,s; DINOv2 \texttt{CLS} similarity between an early window and the final $10$\,s, on the full frame and on the detected subject crop).
}
\label{tab:identity_long}
\vspace{0.5em}

\resizebox{0.9\linewidth}{!}{%
\begin{tabular}{lcccc}
\toprule
\textbf{Configuration} & \textbf{Frame Sim.} & \textbf{Wins} & \textbf{Crop Sim.} & \textbf{Wins} \\
\midrule
Baseline & 0.745 & -- & 0.642 & -- \\
$+$ \textbf{Ours} & \textbf{0.850} & 76\% & \textbf{0.696} & 64\% \\
\bottomrule
\end{tabular}%
}

\end{table}

\subsection{Identity Retention over Extended Horizons}
\label{sec:suppl_identity}

To quantify how long this stability persists, we measure identity retention as the DINOv2~\cite{dinov2} \texttt{CLS}-token cosine similarity of each frame to frame~$0$. The same prompts and frame times are used for both methods, so the \emph{gap} between the curves is the meaningful quantity; the absolute value is below $1$ even for good videos because legitimate scene motion changes the global feature.

\cref{tab:identity} reports the result. Over five minutes the recency baseline decays to $0.61$, while \ourmodel stops decaying after the first minute and holds identity flat to the end ($0.81$ at $1$\,min and at $5$\,min). 

\begin{table}[!ht]
\centering
\small
\setlength{\tabcolsep}{10pt}
\caption{\textbf{Identity retention over time}
($20$ prompts $\times$ $5$ seeds. ``mid''/``end'' are $1$/$5$\,min for the five-minute rows and $30$/$60$\,s for the $60$-second rows). \ourmodel holds identity where the recency baseline drifts.}
\vspace{0.5em}
\label{tab:identity}
\begin{tabular}{llccc}
\toprule
\textbf{Horizon} & \textbf{Method} & \textbf{start} & \textbf{mid} & \textbf{end} \\
\midrule
\multirow{2}{*}{$5$\,min}
    & Baseline       & 1.00 & 0.63 & 0.61 \\
    & \textbf{+ Ours} & 1.00 & \textbf{0.81} & \textbf{0.81} \\
\midrule
\multirow{2}{*}{$60$\,s}
    & Baseline       & 1.00 & 0.70 & 0.62 \\
    & \textbf{+ Ours} & 1.00 & \textbf{0.87} & \textbf{0.88} \\
\bottomrule
\end{tabular}
\end{table}

\subsection{Controlled Reappearance}
\label{sec:suppl_reappear}

The experiments above tests whether appearance remains stable while the subject stays visible, but the motivating use case for persistent memory is more demanding: a subject may disappear for longer than the local attention window and later appear again with the same identity.

A prompt-only reappearance benchmark is difficult to construct reliably with current backbones. We initially wrote prompts with staged timing. For example, the subject exits and returns, the scene opens empty, or a second subject enters later. However, current backbones lack the basic ability to execute such prompt-level temporal instructions: mentioned entities appear immediately and essentially never leave. A controlled reappearance evaluation therefore cannot be constructed by prompting alone.

Instead, we construct the reappearance test directly at inference time. Each video is generated as one continuous rollout of four prompt segments: an empty scene ($3$\,s), the subject present ($6$\,s), the same scene explicitly empty ($7.5$\,s, five times the $1.5$\,s attention window), and the subject re-conditioned ($7.5$\,s).

At every segment boundary the sliding window is cleared while the sink and the bank persist. Each segment is therefore generated from the text prompt and the persistent memory alone. Because the opening scene contains no subject, the subject never enters the sink, and the bank is the only channel that can carry its appearance across the gap.

A Qwen2.5-VL check verifies the intended timeline (subject detected in its two segments and absent during the gap). Identity is scored by the same dual metric as in \cref{sec:suppl_protocols} between the first-appearance and return windows.

\ourmodel clearly improves both measures over the base model (\cref{tab:reappear_ctrl}). The base model re-invents the subject from the prompt alone---often with visibly different appearance---whereas with \ourmodel the returning subject is rendered against its stored first-appearance states, of which several hundred remain resident in the bank at the moment of return in every rollout.

\begin{table}[!ht]
\centering
\caption{
\textbf{Controlled Reappearance}. Subject enters, leaves for a $7.5$\,s gap ($5\times$ the attention window) with the sliding window cleared at every shot boundary, then re-conditioned. 
}
\vspace{0.5em}
\label{tab:reappear_ctrl}
\resizebox{\linewidth}{!}{
\begin{tabular}{lcccc}
\toprule
\textbf{Configuration} & \textbf{Frame Sim.} & \textbf{Wins} & \textbf{Crop Sim.} & \textbf{Wins} \\
\midrule
Baseline & 0.699 & -- & 0.648 & -- \\
$+$ \textbf{Ours} & \textbf{0.752} & 73\% & \textbf{0.688} & 66\% \\
\bottomrule
\end{tabular}
}
\vspace{-1em}
\end{table}

\section{Memory Bank Analysis}
\label{sec:suppl_bank_analysis}

The previous section establishes that persistent appearance memory improves long-horizon identity. We now analyze the mechanism itself: which historical states should be admitted into the bank, and whether the resulting memory remains genuinely time-neutral rather than degenerating into a recency buffer.

\subsection{Admission Rule: What Should Enter the Bank?}
\label{sec:suppl_admission}

\cref{tab:picker} varies only the bank admission rule under the main protocol ($128$ prompts $\times\,5$ seeds).
\textit{Random History} samples entries uniformly from previously observed content, whereas \textit{Recent History} retains the most recently observed entries. \textit{DINOv2 Diversity} embeds all historical candidates with DINOv2 and applies K-means to select representatives near the cluster centers, favoring diversity in semantic feature space. \textit{Flow Novelty (ours)} instead admits candidates according to the novelty of their optical-flow maps, explicitly favoring histories with distinct motion patterns.

Random and recency-based selection provide little benefit: Random History and Recent History improve Dynamic Degree by only $+1.1$ and $+0.7$ over the no-bank baseline ($62.4$/$62.0$ vs.\ $61.3$), while random selection even lowers the Total score ($77.7$ vs.\ $78.3$). DINOv2 Diversity improves the Total to $79.4$ but does not recover motion ($60.6$ Dynamic Degree), suggesting that diversity in semantic feature space is not the signal the bank needs.

In contrast, Flow Novelty raises Dynamic Degree to $71.3$ ($+10.0$)
while preserving consistency, and achieves the best Total score ($79.7$).
These results show that what enters the bank matters:
admission based on appearance novelty is substantially more effective
than retaining arbitrary, recent, or semantically diverse history
for long video generation.

\begin{table}[!bth]
\centering
\scriptsize
\setlength{\tabcolsep}{2.6pt}
\caption{
\textbf{Ablation of Bank Admission Rules}. 
}
\label{tab:picker}
\resizebox{\linewidth}{!}{
\begin{tabular}{lcccccc|ccc}
\toprule
 & \textbf{S.C.} & \textbf{B.C.} & \textbf{M.S.} & \textbf{D.D.} &
\textbf{A.Q.} & \textbf{I.Q.} & \textbf{Q.} & \textbf{Sem.} & \textbf{Total} \\
\midrule
 No Bank (Baseline) & 97.4 & 95.9 & 98.4 & 61.3 & 55.8 & 68.8 & 82.2 & 62.9 & 78.3 \\
\midrule
Random History & 97.1 & 96.9 & 98.5 & 62.4 & 56.4 & 69.2 & 82.4 & 58.7 & 77.7 \\
Recent History & 97.4 & 96.1 & 98.4 & 62.0 & 56.0 & 69.0 & 82.3 & 64.2 & 78.7 \\
DINOv2 Diversity & 97.6 & 96.1 & 98.6 & 60.6 & 57.6 & 69.4 & 82.9 & 65.6 & 79.4 \\
Flow Novelty (Ours) & 97.4 & 96.0 & 98.4 & \textbf{71.3} & 57.5 & 68.8 & 83.3 & 65.4 & \textbf{79.7} \\
\bottomrule
\end{tabular}
}
\end{table}

\subsection{Bank Retention Behavior}
\label{sec:suppl_retention}

To empirically verify the time-neutral retention described in \cref{sec:bank}, we log per-block bank traces over $220$ rollouts. At each step, the bank closely matches the top-$K$ candidates among \emph{all} items observed so far, ranked by their frozen scores, achieving a final-state overlap of $0.99$. The small residual discrepancy is attributable to score ties.

Consequently, eviction removes almost exclusively below-cutoff admissions, predominantly near-zero-score content admitted while the bank is still filling. Moreover, the source-frame ages of the final bank entries are distributed roughly uniformly across the rollout, with a mean age near the midpoint of the video. This behavior is characteristic of time-neutral retention: survival is determined by novelty rather than recency.

\section{Inference Cost}
\label{sec:suppl_speed}

\cref{tab:vbench_long_speed} reports the wall-clock time for generating $1$-, $2$-, and $5$-minute videos on a single GH200 GPU. \ourmodel incurs a roughly constant $1.6\times$ overhead that does not increase with video length. A naive bank refill would scale quadratically with duration and require hundreds of GB of cached history. Instead, we store only novel-token activations and refill bank slots incrementally, reproducing the naive path bit-identically while keeping the per-block cost constant.

\begin{table}[!ht]
\centering
\small
\setlength{\tabcolsep}{5pt}
\caption{
\textbf{Generation Time} for a single video of resolution of $832\times480$ on a 80GB GH200 GPU. \ourmodel adds a $\sim$1.6$\times$ overhead that does not grow with length.
}
\label{tab:vbench_long_speed}
\begin{tabular}{lcccc}
\toprule
\textbf{Length} & \textbf{Frames (blocks)} & \textbf{Baseline} & \textbf{Ours} & \textbf{Overhead} \\
\midrule
$1$ min & \phantom{0}240 (\phantom{0}80) & \phantom{0}185\,s & \phantom{0}288\,s & $1.56\times$ \\
$2$ min & \phantom{0}480 (160) & \phantom{0}268\,s & \phantom{0}450\,s & $1.68\times$ \\
$5$ min & 1200 (400) & 1435\,s & 2239\,s & $1.56\times$ \\
\bottomrule
\end{tabular}
\end{table}

\section{Is the Gain Good Motion? A Camera--Object Decomposition Analysis}
\label{sec:suppl_motion_decomp}

\begin{table}[!ht]
\centering
\small
\setlength{\tabcolsep}{6pt}
\caption{
\textbf{Camera--Object Motion Decomposition}
\ourmodel \textbf{reduces} camera motion while largely preserving object motion, raising the object share. Drift tail counts videos whose mean total flow exceeds $2\times$ the config median.
}
\label{tab:motion_decomp}
\resizebox{\linewidth}{!}{
\begin{tabular}{lccccc}
\toprule
\textbf{Config} &
\textbf{Camera Flow} &
\textbf{Object Flow} &
\textbf{Object Fraction} &
\textbf{Total Flow} &
\textbf{Drift Tail} \\
\midrule
Baseline        & 1.204 & 0.710 & 0.380 & 1.914 & 37 \\
 $+$ \textbf{Ours} & 0.897 & 0.659 & 0.408 & 1.556 & 24 \\
\bottomrule
\end{tabular}
}
\end{table}

A natural concern is that the gain in Dynamic Degree is the cheap kind of motion---global camera panning, which inflates the metric without improving the video. However, it turns out to be the opposite.

For each video, we compute dense optical flow with RAFT on frame pairs sampled every ${\sim}0.5$\,s, fit a global homography with RANSAC, and decompose the flow into a homography-predicted camera component and a residual object-motion component. \cref{tab:motion_decomp} reports the mean magnitudes over all $128$ prompts.

On Self-Forcing, \ourmodel \emph{reduces} the baseline's camera motion ($1.20\!\to\!0.90$) while largely preserving object motion ($0.71\!\to\!0.66$), increasing the fraction of motion attributable to objects from $37\%$ to $42\%$. Thus, \ourmodel does not improve the metric by injecting additional camera motion. Instead, it suppresses the baseline's erratic global drift while retaining most of the subject's motion.

Dynamic Degree measures the fraction of videos whose motion exceeds a genuine-motion threshold rather than the mean flow magnitude, so it can increase even when mean flow decreases. \ourmodel pushes the baseline's near-frozen clips above this threshold while suppressing its runaway-drift outliers. Accordingly, the per-video total-flow deviation decreases from $\pm2.9$ to $\pm1.9$, while the number of clips in the drift tail falls from $37$ to $24$.

\end{document}